\documentclass[eat,twocolumn]{jmlr}

\usepackage{longtable}

\usepackage{booktabs}
\usepackage[load-configurations=version-1]{siunitx} 

\theorembodyfont{\upshape}
\theoremheaderfont{\scshape}
\theorempostheader{:}
\theoremsep{\newline}

\jmlrvolume{}
\firstpageno{1}

\jmlryear{2022}
\jmlrworkshop{Machine Learning for Health (ML4H) 2022}

\title[Algorithmic bias in delirium prediction]{Algorithmic bias in machine learning based delirium prediction}

  \author{\Name{Sandhya Tripathi} \Email{sandhyat@wustl.edu}\\
  \Name{Bradley A Fritz} \Email{bafritz@wustl.edu}\\
  \Name{Michael S Avidan} \Email{avidanm@wustl.edu}\\
  \addr Department of Anesthesiology, Washington University in St Louis,
MO, USA
 \AND
 \Name{Yixin Chen} \Email{ychen25@wustl.edu}\\
 \addr Department of Computer Science and Engineering, Washington
University in St Louis, St Louis, MO, USA
 \AND
  \Name{Christopher R King} \Email{christopherking@wustl.edu}\\
  \addr Department of Anesthesiology, Washington University in St Louis,
MO, USA
}

\begin{document}

\maketitle

\begin{abstract}
Although prediction models for delirium, a commonly occurring condition during general hospitalization or post-surgery, have not gained huge popularity, their algorithmic bias evaluation is crucial due to the existing association between social determinants of health and delirium risk. In this context, using MIMIC-III and another academic hospital dataset, we present some initial experimental evidence showing how sociodemographic features such as sex and race can impact the model performance across subgroups. With this work, our intent is to initiate a discussion about the intersectionality effects of old age, race and socioeconomic factors on the early-stage detection and prevention of delirium using ML.
\end{abstract}

\section{Introduction}
\label{sec:intro}

Delirium, a commonly under-diagnosed and under-treated condition during hospitalization, is characterized by an acute change in awareness, attention, and cognition arising alongside serious illness \citep{american2013diagnostic}. 
Delirium-associated risk factors, symptoms and implications in the elderly population have been widely studied \citep{inouye2014delirium}. 
The delirium prevalence rate can be more than $20\%$ for high-risk post-surgical patients \citep{wilson2020delirium} with up to $50\%$ chances after hip replacement. 
In addition to the impact on patients, the estimated additional healthcare costs associated with delirium exceed $\$44,291$ per patient per year \citep{gou2021one}. 

 Machine learning (ML) based delirium prediction can have several applications in clinical practice. 
First, patients at higher risk of delirium can be prioritized to receive preventative treatments and safety measures to mitigate harmful behaviors should delirium develop.
Accurate targeting is necessary because the most effective treatments are labor-intensive and require high levels of nursing supervision, making them not easily scaled to all at-risk patients.
There are also tradeoffs in patient quality of life with excessively aggressive delirium prevention. 
For example, some anti-anxiety medications (benzodiazepines) and sleep aides are thought to increase delirium risk, but unnecessarily removing these medications from lower-risk patients deprives them of their beneficial effects.

 An important concern while using ML for clinical decision support is algorithmic bias, defined as an ML model's capability to reinforce existing biases against certain population groups (such as certain categories of age, sex, and race commonly referred to as protected or sensitive attributes). 
 With a long-standing discussion on the relation between social determinants of health (SDOH) and delirium risk \citep{khan2016relationshipNoeffect2016,PMCID:PMC8682256HealthdisparitiesAbstract},  there is a pressing need to evaluate delirium prediction models for algorithmic bias and
 make sure that the models do not increase the disparities in delirium prevention. 
 This evaluation for algorithmic bias becomes more crucial because some of the currently available ML-based delirium prediction models use protected attributes such as race, insurance status, and language when these features could be the root cause of dissimilarities during delirium assessment.
 Also, as delirium prediction can also be used to enrich patient samples in research of novel biomarkers, prophylaxis, and treatment, models which are not accurate in sub-populations would then deprive patients of the opportunity to participate in such research and produce studies which do not include all relevant demographics.
 Presently, the research on the subject of bias in ML-based delirium prediction has been limited by sample size and algorithm complexity \citep{castro2021development}.
 
 \textbf{Contributions:} With our study, we 1) present evidence that there could be algorithmic bias in delirium prediction, 2) recommend users to evaluate the prediction models for algorithmic bias and  check if this bias can be reduced by simple methods such as removing the protected attribute or accounting for confounders via Propensity Score Matching.

\section{Related work}
There have been some disagreeing viewpoints among researchers when studying the effect of demographic factors and SDOH on delirium risk \citep{khan2016relationshipNoeffect2016,PMCID:PMC8682256HealthdisparitiesAbstract}.
As suggested by authors in \cite{arias2022framework}, solutions such as the SDOH framework for delirium can capture the challenges and vulnerabilities unique to older adults. Two separate studies investigating delirium risk in medical ICU patients in a Dutch region \citep{wu2021socialDutch}, and surgical ICU patients in the US \citep{khan2016relationshipNoeffect2016} found no significant effect of race or SDOH on delirium risk. In contrast, \cite{PMCID:PMC8682256HealthdisparitiesAbstract} report evidence of a substantial difference in delirium risk between non-Hispanic White and Black patients in the (71-80) age groups. Researchers studying dementia and long-term cognitive impairment (LTCI) discovered that delirium severity was higher in African American participants \citep{boltz2021disparities}. Further, the socioeconomic and clinical risk factors, such as race, education, hospital type, and delirium duration, were linked to worse LTCI \citep{haddad2020socioeconomic}. Health disparities that accentuated during the COVID-19 pandemic require us to be more proactive in terms of screenings for delirium, a well-known complication of respiratory illness in older adults \citep{o2020deliriumCovidEditorial}.

A recent survey by \cite{chua2021predictionReview2021} presents the various ML models developed in different hospital settings. The penetration of ML in delirium prediction is still in the initial stages as the models are elementary. ML-based delirium prediction can be used for patients during post-operative stay \citep{racine2021machine, bishara2022postoperative}, ICU stay \citep{hur2021machine}, general hospitalization \citep{ocagli2021machine}, and in the emergency department \citep{lee2021machine}. 
While these models achieve satisfactory prediction performance, they overlook the possibly harmful inclusion of highly predictive attributes such as insurance status and language in model building as also noted by \cite{chua2021predictionReview2021}.  
A small multicenter study by \cite{castro2021development} does a surface-level evaluation of algorithmic bias in ML-based delirium prediction for COVID-19 patients at the time of admission using an l1-penalized regression algorithm. Their analyses 
 shows that the model does not suffer from algorithmic bias. However, as acknowledged by the authors, their labels could be erroneous; their dataset is comparatively small, and mixes ICU and ward patients in very different contexts. In contrast, both of our datasets are large, ICU based, and contain expert-labelled delirium outcomes. Unlike \cite{castro2021development},
our work is the first to show that there could be algorithmic bias in the predictions of a Random Forest model. Since we performed this analysis before deploying the above model, we have the option of modifying the model or making the user aware of these biases. 

\section{Evaluating algorithmic bias}
An algorithm suffers from disparate mistreatment \citep{zafar2017fairnessWWW} if the algorithm's performance differs for otherwise similar people at different levels of a sensitive feature. 
We use evaluation metrics such as Positive Predictive Value (PPV), Sensitivity and AUROC to report disparate mistreatment.

We also evaluate the classifiers on a \textit{propensity score matched} (PSM) sample so that the distribution of $\mathbf{X}$ is similar between ``treated'' ($\mathbf{Z}_i = 1$) and ``untreated'' ($\mathbf{Z}_i = 0$) subjects \citep{austin2011introduction}. Here $\mathbf{Z}$ and $\mathbf{X}$ denote sensitive and baseline covariates (features other than sensitive features).
Hence, in order to compare only individuals with similar covariates, we calculate $e_i = P(\mathbf{Z}_i =1|\mathbf{X}_i)$ and 1-to-1 match individuals with $z=1$ to individuals with similar $e$ and $z=0$. To estimate the propensity score, we use a logistic regression model with $\mathbf{Z}$ being regressed on observed baseline features ($\mathbf{X}$).

\section{Experimental setup}

In our experiments, we consider two datasets: ACTFAST-Epic dataset and MIMIC-III \citep{johnson2016mimic}. ACTFAST-Epic dataset is electronic health data recorded prior to surgical procedures between November 2018 and Feb 2021 at Barnes-Jewish Hospital (St Louis, MO, USA). Access to the data was approved by the Human Research Protection Office at Washington University in St Louis, USA with a waiver of informed consent (IRB number $201607122$). We used CAM-ICU assessments \citep{ely2001evaluation} in ACTFAST-Epic dataset and ICD-9 codes in MIMIC-III  to assign delirium labels; the  patient demographics are available in Table \ref{tab:actfastEpicDems}. For MIMIC-III, we followed the pre-processing and model training steps as in \cite{coombes2021novel}. After preprocessing, there were a total of $12699$ patients in ACTFAST-Epic and $48435$ patients in the MIMIC-III dataset. For each subgroup of a protected attribute, we report the difference in metric from the mean metric value of the protected attribute and average it to perform bootstrapped t-testfor comparing the metrics. We trained Random Forest classifiers on both ACTFAST-Epic and MIMIC-III datasets. Additional details on datasets and model training are available in Appendix \ref{apd:first}. 

Our experiments are intended to investigate the following hypotheses: 0) there could be performance differences across subgroups measured using different metrics and these differences might (not) be eliminated by either 1) removing the protected variable or 2) accounting for confounding variables (using PSM). 

For compactness, we choose AUROC as the main performance metric.
Table \ref{tab:AUROC-summary} presents the gap between the maximum difference from the average and the minimum difference from the average of the AUROC metric across the levels/subgroups of a protected attribute. The difference was taken post-averaging across 150 bootstraps of the dataset.

\section{Results and Discussion}

Table \ref{tab:actfastEpicPerf} reports the differences in subgroups and that propensity score matching (PSM) can reduce these differences. We also check the effect of removing the protected variable and observe that the performance doesn't change (Table \ref{tab:EpicWOSesnPerf} in Appendix \ref{apd:first}).

As shown in Table \ref{tab:actfastEpicPerf}, for ACTFAST-Epic dataset, the performance differences change modestly on the propensity-matched sample across all the metrics. 
This is due to differences in the procedures and comorbidities included in the propensity score. 
Notably, there are no other SDOHs included in the propensity score which could be acting as surrogates for the race variable.
Differential missing data across subgroups could also contribute to the drop in apparent differential mistreatment when propensity matching, as matching conditional on data may exclude some patients with significant gaps in their records (and worse classifier performance).
For ACTFAST-Epic, since removing the sensitive variable from the training dataset didn't reduce the performance differences, the sensitive attributes could still affect the outcome via other confounding variables.  As an example, height can be a strong confounder in the absence of sex.
We also measured the False negative rate (FNR) and False positive rate (FPR) of the classifiers learnt on ACTFAST-Epic dataset. We noticed that there are higher discrepancies in the FNR values (negative value implies advantage as lower mistakes) meaning that the patients who were delirious and could have benefitted from the intervention were missed. Full details on FNR and FPR are presented in Table \ref{tab:Epic_mistakes}, Appendix \ref{apd:first}. We can also obtain confidence intervals on AUROC using the bootstraps as reported in Table \ref{tab:Epic-auroc-std} for the ACTFAST-Epic dataset.

Table \ref{tab:mimicPerf} reports that adding the sensitive variable  to the feature set (model M2) can change the results in terms of subgroups' performance and hence answers to hypothesis 2. We only included AUROC in the main paper as the others showed a similar pattern of change in performance; other metrics are included in Table \ref{tab:MIMIC-withOutsensM1} and \ref{tab:MIMIC-withsensM2} in Appendix \ref{apd:first}.

Unlike some of the existing delirium prediction models, \cite{coombes2021novel} do not include the protected variables in the model training on MIMIC-III dataset. The subgroup performance of a Random Forest model (\textbf{M1}) on such a feature set is presented in columns 4 and 5 of Table \ref{tab:mimicPerf}. To demonstrate the adverse effect of including the protected attributes, we also trained another model (\textbf{M2}) that used protected attributes in addition to the predictors of (\textbf{M1}) and report the corresponding AUROC in the last two columns of Table \ref{tab:mimicPerf}. Although the overall performance of the two models is the same, the differences between the subgroup and group average of (\textbf{M2}) are statistically significant. 


 As a summarizing result for the two hypotheses mentioned earlier, we are interested in two kinds of comparisons from Table \ref{tab:AUROC-summary}. For a fixed model, we expect that the after-matching values are the same or smaller. Across models, we compare the before-matching/after-matching values to see if removing the protected attribute brings any improvement. Matching reduces the discrepancy in most cases. Removing the protected variable could be helpful in some cases as seen in the case of MIMIC-III. We would like to caution that even though these summaries are easier, they lose information on which subgroup is affected. They could also be affected by the skewness in the subgroup sample size.

Our work is limited by the use of only the top two or three commonly occurring subgroups in protected features such as sex and race. 
To some extent, delirium assessment has a subjective part to it, so extending the protected attribute set to include language and insurance for evaluation of the models can provide insights helpful in understanding the intersectionality of various SDOH in the context of delirium prediction. Another necessary step is the evaluation of the delirium prediction models to reflect the interaction of two protected attributes.

From a broader societal perspective, these studies would benefit the areas where SDOHs affect data used in medical risk prediction models.
Designing minimally biased EHR-based risk stratification for delirium prediction can guide personalized post-operative delirium prevention and reduction strategies without disadvantaging  certain socioeconomic groups.


\begin{table}[]
\floatconts{tab:actfastEpicDems}
{\caption{Protected attribute distribution in ACTFAST-Epic test dataset. Delirium labels were extracted using the CAM-ICU tool.}}
{\scriptsize
\begin{tabular}{|l|c|c|c|}
\hline
\textbf{\begin{tabular}[c]{@{}l@{}}Patient \\ groups\end{tabular}} & \textbf{Total} & \textbf{\begin{tabular}[c]{@{}c@{}}Delirium\\ pos\end{tabular}} & \textbf{\begin{tabular}[c]{@{}c@{}}Delirium\\ neg\end{tabular}} \\ \hline
\textbf{Male} & 1460 & 662 & 798 \\ \hline
\textbf{Female} & 2352 & 1125 & 1227 \\ \hline
\textbf{Black} & 971 & 524 & 447 \\ \hline
\textbf{White} & 2747 & 1220 & 1527 \\ \hline
\textbf{Race NA} & 94 & 43 & 51 \\ \hline
\textbf{Age {[}14 - 52)} & 1247 & 546 & 701 \\ \hline
\textbf{Age {[}52 - 67)} & 1278 & 602 & 676 \\ \hline
\textbf{Age {[}67 - 102{]}} & 1287 & 639 & 648 \\ \hline
\textbf{Total} & 3812 & 1787 & 2025 \\ \hline
\end{tabular}}
\end{table}

\begin{table*}[h!]
\floatconts{tab:PerfDiff}
  {\caption{Performance difference from the group average for (a) the ACTFAST-Epic dataset  and (b) MIMIC-III dataset using a Random Forest classifier on a bootstrapped (150 samples) test set. \textasciicircum{}Multiple values imply results when matched to patients in each group in this category in the same order as in column `Patient group'. Subtable (b) compares AUROC for \textbf{M1 (without protected attributes)} and \textbf{M2 (with protected attributes)} models along with reporting the distribution of protected attributes conditional on the outcome value in the test set of MIMIC-III. Statistical significance at $\alpha = 0.05$ is symbolized by *.}}
 {%
   \subtable[][c]{%
     \label{tab:actfastEpicPerf}%
     {\scriptsize \begin{tabular}{|l|c|c|c|c|c|c|}
\hline
\textbf{\begin{tabular}[c]{@{}l@{}}Patient \\ group\\ ACTFAST-Epic\end{tabular}} & \textbf{\begin{tabular}[c]{@{}c@{}}Grp - Avg\\  PPV\end{tabular}} & \textbf{\begin{tabular}[c]{@{}c@{}}Grp - Avg \\ PPV (PSM\textasciicircum{})\end{tabular}} & \textbf{\begin{tabular}[c]{@{}c@{}}Grp-Avg\\ Sens\end{tabular}} & \textbf{\begin{tabular}[c]{@{}c@{}}Grp -Avg\\ Sens (PSM)\end{tabular}} & \textbf{\begin{tabular}[c]{@{}c@{}}Grp - Avg \\ AUROC\end{tabular}} & \textbf{\begin{tabular}[c]{@{}c@{}}Grp - Avg \\ AUROC \\ (PSM)\end{tabular}} \\ 
\hline
\textbf{Male} & -0.01* & -0.02* & -0.02* & 0.0 & -0.01* & -0.01* \\ \hline
\textbf{Female} & 0.01* & 0.02* & 0.02* & 0.0 & 0.01* & 0.01* \\ \hline
\textbf{White} & -0.04* & -0.02* & -0.04* & 0.05* & -0.01* & 0.02* \\ \hline
\textbf{Black} & 0.03* & 0.02* & 0.02* & -0.05* & 0.01* & -0.02* \\ \hline
\textbf{Age {[}14 - 52)} & -0.02* & -0.04* & 0.05* & 0.03* & 0.02* & 0.01* \\ \hline
\textbf{Age {[}52 - 67)} & 0.0 & {[}0.04*, -0.02*{]} & -0.01* & {[}-0.03*, 0.01*{]} & -0.01* & {[}-0.01*, '0.0'{]} \\ \hline
\textbf{Age {[}67 - 102{]}} & 0.02* & 0.02* & -0.04* & -0.01* & -0.01* & 0.0 \\ \hline
\textbf{Overall Avg} & 0.76 (PPV) & - & 0.77 (Sens) & - & 0.85 (AUROC) & - \\ \hline
\end{tabular}}
   }\qquad
   \subtable[][c]{%
     \label{tab:mimicPerf}%
     {\scriptsize \begin{tabular}{|l|c|c|c|c|c|c|}
\hline
\textbf{\begin{tabular}[c]{@{}l@{}}Patient \\ groups \\ MIMIC-III\end{tabular}} & \textbf{\begin{tabular}[c]{@{}c@{}}Delirium \\ Pos\end{tabular}} & \textbf{\begin{tabular}[c]{@{}c@{}}Delirium \\ Neg\end{tabular}} & \textbf{\begin{tabular}[c]{@{}c@{}}Grp - Avg \\ AUROC \\ M1\end{tabular}} & \textbf{\begin{tabular}[c]{@{}c@{}}Grp - Avg \\ AUROC \\ M1 (PSM)\end{tabular}} & \textbf{\begin{tabular}[c]{@{}c@{}}Grp - Avg \\ AUROC \\ M2\end{tabular}} & \textbf{\begin{tabular}[c]{@{}c@{}}Grp - Avg \\ AUROC \\ M2 (PSM)\end{tabular}} \\ \hline
\textbf{Male} & 565 & 6066 & 0.01* & 0.01* & 0.01* & 0.02* \\ \hline
\textbf{Female} & 413 & 4896 & -0.01* & -0.01* & -0.01* & -0.02* \\ \hline
\textbf{Hispanic} & 27 & 338 & 0.0 & {[}0.01*, -0.0{]} & 0.02* & {[}0.03*, 0.01*{]} \\ \hline
\textbf{White} & 845 & 9411 & 0.0 & {[}-0.0*, 0.0{]} & -0.01* & {[}-0.01*, -0.01*{]} \\ \hline
\textbf{Black} &  106 & 1213 & 0.01* & {[}0.0*, -0.01*{]} & -0.01* & {[}0.01*, -0.03*{]} \\ \hline
\textbf{Age {[}18 - 57)} & 269 & 3709 & -0.01* & 0.0 & -0.01* & 0.01* \\ \hline
\textbf{Age {[}57 - 74)} & 296 & 3685 & 0.01* & {[}0.0, 0.0{]} & 0.0 & {[}-0.01*, -0.01*{]} \\ \hline
\textbf{Age {[}74 - 90{]}} & 413 & 3568 & 0.01* & 0.0 & 0.01* & 0.01* \\ \hline
\textbf{Overall Avg} & 978 & 10962 & 0.87 (AUROC) & - & 0.87 (AUROC) & - \\ \hline
\end{tabular}}
   }
 }
\end{table*}

\begin{table*}[!h]
\floatconts{tab:AUROC-summary}
{\caption{{Summarizing metric (on AUROC) between models trained on datasets with and without protected attributes with further comparison using propensity matching.}}}
{\scriptsize
\begin{tabular}{|c|cccc|}
\hline
 & \multicolumn{4}{c|}{\textbf{Max discrepancy in AUROC across subgroups for each protected attribute}} \\ \hline
\textbf{\begin{tabular}[c]{@{}c@{}}Protected \\ attribute\end{tabular}} & \multicolumn{2}{c|}{\textbf{Model with protected attribute}} & \multicolumn{2}{c|}{\textbf{Model without protected attribute}} \\ \hline
 & \multicolumn{1}{c|}{\textbf{Before matching}} & \multicolumn{1}{c|}{\textbf{After Matching}} & \multicolumn{1}{c|}{\textbf{Before Matching}} & \textbf{After Matching} \\ \hline
 & \multicolumn{4}{c|}{\textbf{MIMIC-III dataset}} \\ \hline
\textbf{Race} & \multicolumn{1}{c|}{0.03} & \multicolumn{1}{c|}{0.04} & \multicolumn{1}{c|}{0.01} & 0.01 \\ \hline
\textbf{Sex} & \multicolumn{1}{c|}{0.02} & \multicolumn{1}{c|}{0.04} & \multicolumn{1}{c|}{0.02} & 0.02 \\ \hline
\textbf{Age} & \multicolumn{1}{c|}{0.02} & \multicolumn{1}{c|}{0.02} & \multicolumn{1}{c|}{0.02} & 0 \\ \hline
 & \multicolumn{4}{c|}{\textbf{ACTFAST-Epic dataset}} \\ \hline
\textbf{Race} & \multicolumn{1}{c|}{0.02} & \multicolumn{1}{c|}{0.04} & \multicolumn{1}{c|}{0.02} & 0.02 \\ \hline
\textbf{Sex} & \multicolumn{1}{c|}{0.02} & \multicolumn{1}{c|}{0.02} & \multicolumn{1}{c|}{0.02} & 0.02 \\ \hline
\textbf{Age} & \multicolumn{1}{c|}{0.03} & \multicolumn{1}{c|}{0.01} & \multicolumn{1}{c|}{0.04} & 0 \\ \hline
\end{tabular}}
\end{table*}

\section{Conclusion}
We present our initial investigations to explicitly evaluate ML-based delirium prediction models for algorithmic bias. 
Non-representative dataset samples and biases during manual assessment (\cite{yang2008participation, jones2006does}) add layers of complexity to learning effectively. Therefore, we should be aware of the challenges before (evaluating subgroup performance), during (checking  model applicability) and after (making the user aware of the model's pros and cons in terms of bias and applicability) the deployment of prediction models.



\bibliography{pmlr-sample}

\appendix
\section{Additional experimental details}\label{apd:first}

We consider two datasets: 1) Electronic health data (ACTFAST-Epic dataset) recorded prior to surgical procedures between November 2018 and Feb 2021 Barnes-Jewish Hospital (St Louis, MO, USA). Access to the data was approved by the Human Research Protection Office at Washington University in St Louis, USA with a waiver of informed consent (IRB number $201607122$) 2) MIMIC-III dataset \citep{johnson2016mimic}. For the ACTFAST-Epic dataset, CAM-ICU assessments \citep{ely2001evaluation} were used to assign delirium labels and data recorded during preop assessment was used as features.  A complete list of ACTFAST-Epic features used to train the model included a bag of word representations of preop notes (500 dimensional) and preop records (185 dimensional) as presented in Table \ref{tab:EpicFeatures}. For MIMIC-III, delirium labels were extracted using the ICD-9 codes and all the pre-processing and model training steps were performed as by \cite{coombes2021novel}. 
After preprocessing, there were a total of $12699$ patients in ACTFAST-Epic and $48435$ patients in the MIMIC-III dataset.

We trained Logistic Regression, Decision Tree, Random Forest, Gradient Boosted Trees, and DNN models to learn the classifiers on both ACTFAST-Epic and MIMIC-III. We chose Random Forest for reporting as it performed the best in the group of algorithms. The same analysis can be conducted for other algorithms too. Each dataset was split into the train (70\%) and test (30\%) subsets while maintaining the same delirium outcome rate in the splits. We performed the computation on 150 bootstrapped samples of the test split in each dataset. For each subgroup (PSM subgroup), we report the difference from the average metric under consideration. While computing PPV and Sensitivity, we used Youden's index \citep{youden1950index} to identify the classifier's decision boundary. The outcome distribution in the two datasets is different as ACTFAST-Epic is balanced while MIMIC-III is not. Hence, for the MIMIC-III dataset, we only report AUROC in the main paper. The calibration plots for the models on which the classifiers were trained are presented in Figure \ref{fig:calibrationplots}.

We filtered the subgroups for Race/Ethnicity in the datasets based on the data availability in each subgroup and hence reported the subgroup performance for Black, White, and Hispanic (in MIMIC-III) and ``not available/unknown'' (ACTFAST-Epic) groups. We required that at least 100 samples be available after propensity score matching, to compute the final metric. 

\paragraph{Code availability}
We provide the python code that was used to obtain the results for both datasets at \url{https://github.com/sandhyat/AlgorithmicBias_Delirium_ML4H22022}. ACTFAST-Epic dataset is not publicly available. For MIMIC-III, we also provide psql queries used to extract the initial dataset following \cite{coombes2021novel}. 


\begin{table*}[]
\centering
\floatconts{tab:EpicFeatures}
{\caption{Predictors used to train the Random forest model in ACTFAST-Epic dataset.}}
{\tiny
\begin{tabular}{|l|l|l|l|l|}
\hline
\multicolumn{1}{|c|}{\textbf{Demographics}} & \multicolumn{1}{c|}{\textbf{Comorbidities}} & \multicolumn{1}{c|}{\textbf{Comorbidities}} & \multicolumn{1}{c|}{\textbf{Preop Lab Test}} & \multicolumn{1}{c|}{\textbf{Preop Lab Test}} \\ \hline
Sex & Anemia & pastTransplant & POIKILOCYTOSIS & TROPONIN I \\ \hline
emergency & Asthma & LVEF & HCG,URINE, POC & HEMOGLOBIN A1C \\ \hline
SurgService\_Name & CAD & CancerCurrent & CLARITY, URINE & HEMATOCRIT \\ \hline
ETHNICITY & CHF & PacemakerICD & COLOR, URINE & PCO2, ARTERIAL \\ \hline
RACE & CKD & Stroke & \begin{tabular}[c]{@{}l@{}}LEUKOCYTE  ESTERASE, \\  URINE\end{tabular} & LIPASE \\ \hline
age & COPD & PNA & \begin{tabular}[c]{@{}l@{}}RED BLOOD CELLS,\\  URINE\end{tabular} & \begin{tabular}[c]{@{}l@{}}MEAN PLATELET \\ VOLUME\end{tabular} \\ \hline
Temp & CancerHx & delirium\_history & ANISOCYTOSIS & PCO2, VENOUS \\ \hline
TOBACCO\_USE & Chronic\_Pain & MentalHistory\_adhd & HYALINE CAST & \begin{tabular}[c]{@{}l@{}}IRON BINDING  \\ CAPACITY, TOTAL\end{tabular} \\ \hline
HEIGHT\_IN\_INCHES & Cirrhosis & MentalHistory\_other & \begin{tabular}[c]{@{}l@{}}COVID-19  \\ CORONAVIRUS RNA\end{tabular} & \begin{tabular}[c]{@{}l@{}}SPECIFIC GRAVITY, \\ URINE\end{tabular} \\ \hline
WEIGHT\_IN\_KG & DVT\_PE & DVT & URINE NITRITE & \begin{tabular}[c]{@{}l@{}}N-TERMINAL PRO \\ B-TYPE  NATURETIC \\ PEPTIDE\end{tabular} \\ \hline
 & \begin{tabular}[c]{@{}l@{}}Dementia\_Mild \\ CognitiveImpairment\end{tabular} & PE & \begin{tabular}[c]{@{}l@{}}GLUCOSE, URINE,\\  QUALITATIVE\end{tabular} & \begin{tabular}[c]{@{}l@{}}ASPARTATE \\ AMINOTRANSFERASE\end{tabular} \\ \hline
\multicolumn{1}{|c|}{\textbf{\begin{tabular}[c]{@{}c@{}}Clinician \\ Assessment\end{tabular}}} & Diabetes & cancerStatus & WHITE BLOOD CELLS,  URINE & MAGNESIUM \\ \hline
 & Dialysis & CHF\_class & BACTERIA, URINE & BASOPHIL ABSOLUTE \\ \hline
wheezing & GERD & DHF & \begin{tabular}[c]{@{}l@{}}EPITHELIAL CELLS,  \\ SQUAMOUS, URINE\end{tabular} & WHITE BLOOD  CELL COUNT \\ \hline
Pulse & HTN & DM\_Type1 & HEPATITIS C AB & \begin{tabular}[c]{@{}l@{}}RED CELL  \\ DISTRIBUTION  WIDTH SD\end{tabular} \\ \hline
Resp & LiverDisease & ESRD & \begin{tabular}[c]{@{}l@{}}PROTEIN, URINE \\  QUALITATIVE\end{tabular} & PHOSPHORUS, PLASMA \\ \hline
\begin{tabular}[c]{@{}l@{}}Short Blessed\\  Total Score\end{tabular} & MI & MR & BILIRUBIN, DIRECT & VANCOMYCIN TROUGH \\ \hline
History of Falling & \begin{tabular}[c]{@{}l@{}}MentalHistory\\ \_ anxiety\end{tabular} & DiastolicFunction & \begin{tabular}[c]{@{}l@{}}IMMATURE GRANULOCYTE  \\ ABSOLUTE\end{tabular} & TRIGLYCERIDES \\ \hline
Pain Score & \begin{tabular}[c]{@{}l@{}}MentalHistory\\ \_bipolar\end{tabular} & DyspneaFreq & FEV1Percent & BILIRUBIN, TOTAL \\ \hline
poorDentition & \begin{tabular}[c]{@{}l@{}}MentalHistory\\ \_depression\end{tabular} &  & IONIZED CALCIUM & CHLORIDE \\ \hline
ASA & \begin{tabular}[c]{@{}l@{}}MentalHistory\\ \_schizophrenia\end{tabular} & \multicolumn{1}{c|}{\textbf{\begin{tabular}[c]{@{}c@{}}In-hospital \\ attributes\end{tabular}}} & LYMPHOCYTE ABSOLUTE & FIBRINOGEN, CLAUSS \\ \hline
Gait/Transferring & MS &  & URINE UROBILINOGEN & PO2, VENOUS \\ \hline
NutritionRisk & OSA & preop\_del & FERRITIN & C REACTIVE PROTEIN \\ \hline
Barthel index score & OtherRhythm & preop\_ICU & HEMOGLOBIN & RED BLOOD CELL \\ \hline
AD8 Dementia Score & OutpatientInsulin & preop\_vent\_b & CALCIUM & LACTATE \\ \hline
StopBANG & PAD & PlannedAnesthesia & NEUTROPHIL ABSOLUTE & \begin{tabular}[c]{@{}l@{}}THYROID STIMULATING \\  HORMONE\end{tabular} \\ \hline
Mental Status & PHTN & pre\_aki\_status & ALBUMIN & URINE BLOOD \\ \hline
FunctionalCapacity & Pacemaker & plannedDispo & CHOLESTEROL, TOTAL & SO2 (MEAS) ARTERIAL \\ \hline
morse & ReplacedValve & preop\_los & URINE KETONES & inr \\ \hline
SpO2 & Seizures & \begin{tabular}[c]{@{}l@{}}BOW\_NA  \\ (Absence of \\ procedure text)\end{tabular} & EOSINOPHIL ABSOLUTE & \begin{tabular}[c]{@{}l@{}}PROTEIN, TOTAL,  \\ PLASMA\end{tabular} \\ \hline
SBP & Thyroid\_disease & noROS & blood urea nitrogen & PH, VENOUS \\ \hline
DBP & activeInfection & \begin{tabular}[c]{@{}l@{}}hasBlock\\  (kind of anesthesia)\end{tabular} & CREATININE & MONOCYTE ABSOLUTE \\ \hline
Ambulatory Aids & AR &  & IRON, TOTAL & CHOLESTEROL, HDL \\ \hline
Secondary Diagnosis & on\_o2 & \multicolumn{1}{c|}{\textbf{\begin{tabular}[c]{@{}c@{}}Preop Lab\\  Test\end{tabular}}} & PLATELET & \begin{tabular}[c]{@{}l@{}}ALANINE\\  AMINOTRANSFERASE\end{tabular} \\ \hline
 & pancreatitis &  & \begin{tabular}[c]{@{}l@{}}ACTIVATED PARTIAL  \\ THROMBOPLASTIN TIME\end{tabular} & POTASSIUM, PLASMA \\ \hline
\multicolumn{1}{|c|}{\textbf{\begin{tabular}[c]{@{}c@{}}Preop Lab\\  Test\end{tabular}}} & pastDialysis & \begin{tabular}[c]{@{}l@{}}RBC \\ MORPHOLOGY\end{tabular} & \begin{tabular}[c]{@{}l@{}}NUCLEATED RBC ABS, \\ AUTO\end{tabular} & PH, URINE \\ \hline
Coombs & AF & ABORH PAT INTERP & GLUCOSE & \begin{tabular}[c]{@{}l@{}}ALKALINE\\  PHOSPHATASE\end{tabular} \\ \hline
\begin{tabular}[c]{@{}l@{}}CREATINE KINASE,\\  TOTAL\end{tabular} & TR & \begin{tabular}[c]{@{}l@{}}ERYTHROCYTE  \\ SEDIMENTATION RATE\end{tabular} & CO2, TOTAL & SODIUM \\ \hline
\begin{tabular}[c]{@{}l@{}}VITAMIN D, \\ 25-HYDROXY\end{tabular} & AS & \begin{tabular}[c]{@{}l@{}}TRANSFERRIN \\ SATURATION\end{tabular} & PO2 ARTERIAL POC & PH, ARTERIAL \\ \hline
\end{tabular}}
\end{table*}

\begin{figure*}[htbp]
\floatconts
  {fig:calibrationplots}
  {\caption{Calibration plot for RF models reported in the main paper.}}
  {%
    \subfigure[ACTFAST-Epic dataset (Table 2a)]{\label{fig:Epicwithsense}%
      \includegraphics[width=0.6\linewidth]{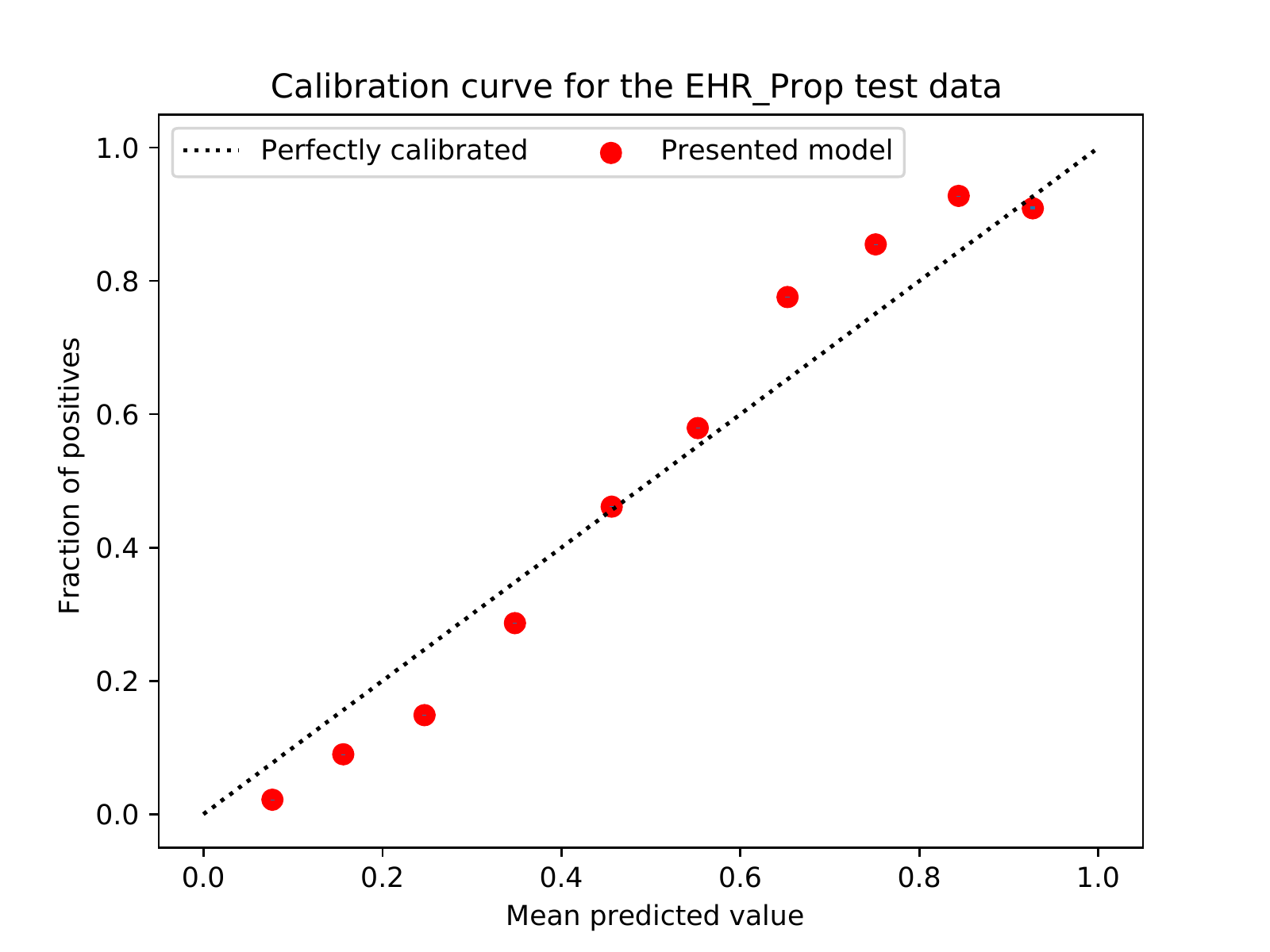}}%
    \qquad
    \subfigure[MIMIC-III dataset Model M1 (Table 2b)]{\label{fig:MIMICmodelM1}%
      \includegraphics[width=0.6\linewidth]{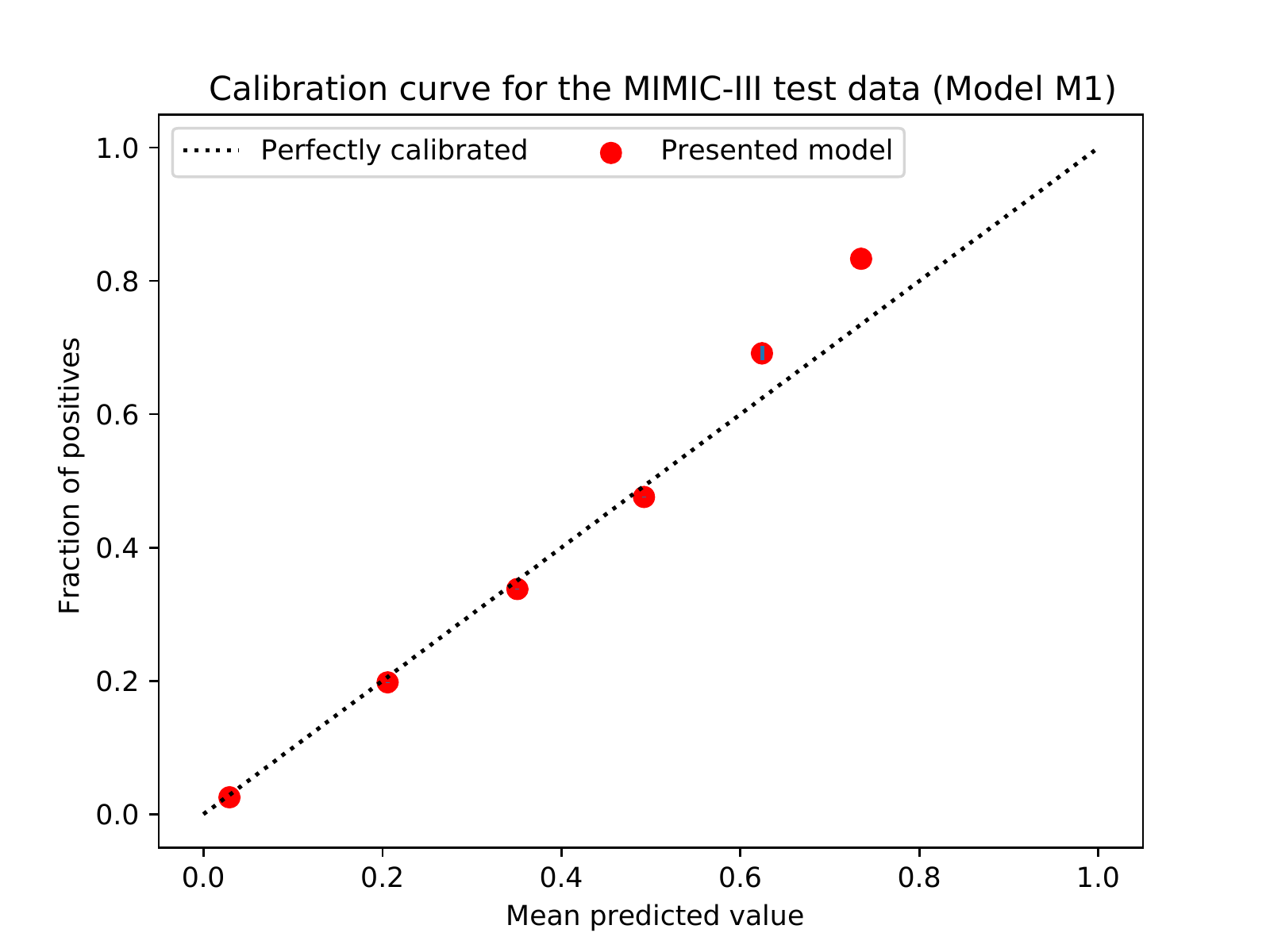}}
    \qquad
    \subfigure[MIMIC-III dataset Model M2 (Table 2b)]{\label{fig:MIMICmodelM2}%
      \includegraphics[width=0.6\linewidth]{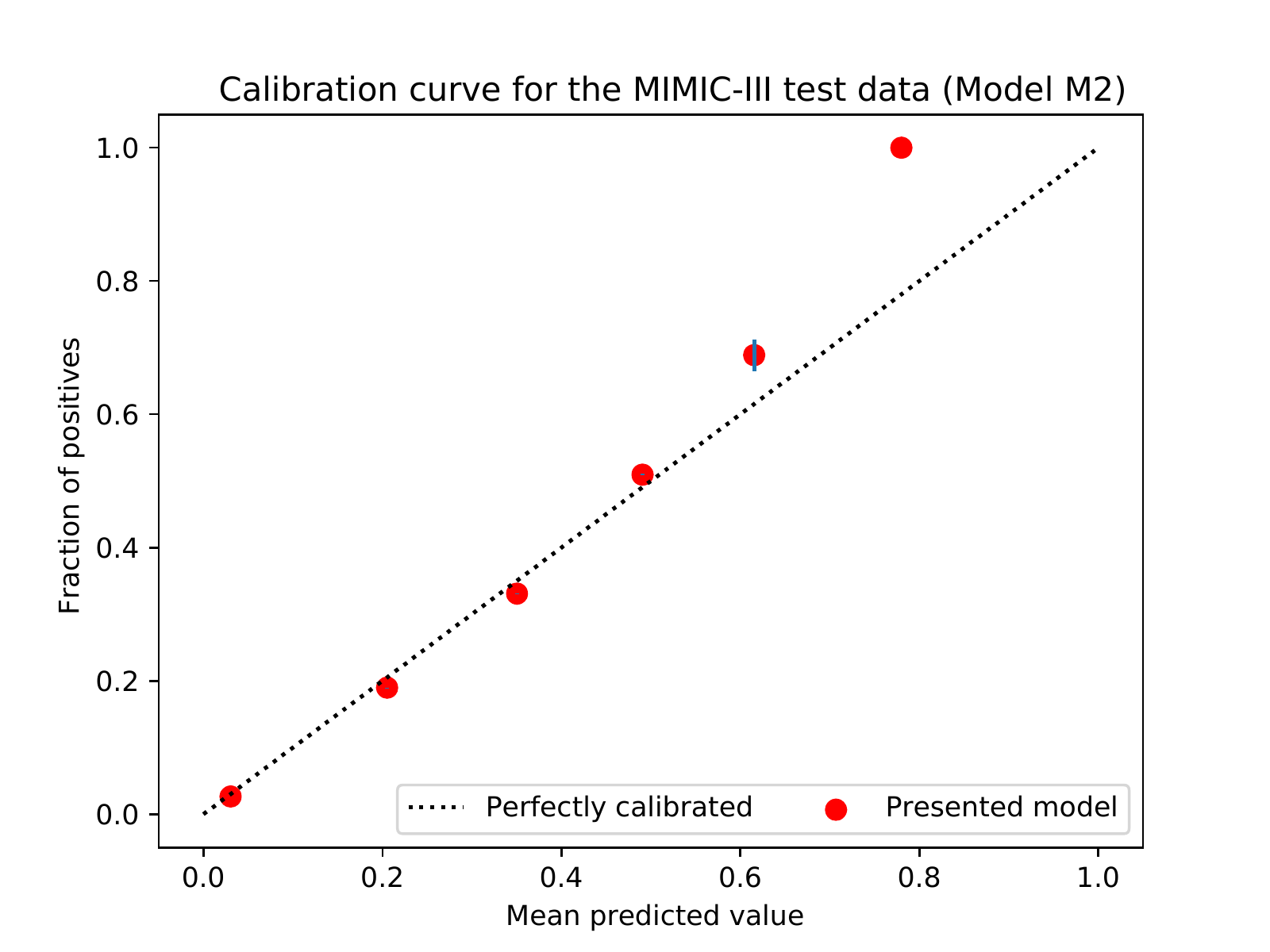}}
  }
\end{figure*}

\begin{table*}[h!]
\floatconts{tab:EpicWOSesnPerf}
  {\caption{Performance difference from the group average for the ACTFAST-Epic dataset using a Random Forest classifier on a bootstrapped (150 samples) test set when the sensitive features (age, sex, and race) \textbf{were not} included in the training dataset. \textasciicircum{}Multiple values imply results when matched to patients in each group in this category in the same order as in column `Patient group'.}}
 {
     {\scriptsize \begin{tabular}{|l|c|c|c|c|c|c|}
\hline
\textbf{\begin{tabular}[c]{@{}l@{}}Patient \\ group\\ ACTFAST-Epic\end{tabular}} & \textbf{\begin{tabular}[c]{@{}c@{}}Grp - Avg\\  PPV\end{tabular}} & \textbf{\begin{tabular}[c]{@{}c@{}}Grp - Avg \\ PPV (PSM\textasciicircum{})\end{tabular}} & \textbf{\begin{tabular}[c]{@{}c@{}}Grp-Avg\\ Sens\end{tabular}} & \textbf{\begin{tabular}[c]{@{}c@{}}Grp -Avg\\ Sens (PSM)\end{tabular}} & \textbf{\begin{tabular}[c]{@{}c@{}}Grp - Avg \\ AUROC\end{tabular}} & \textbf{\begin{tabular}[c]{@{}c@{}}Grp - Avg \\ AUROC \\ (PSM)\end{tabular}} \\ 
\hline
\textbf{Male} & -0.01* & -0.03* & -0.01* & 0.02* & -0.01* & -0.01* \\ \hline
\textbf{Female} & 0.01* & 0.03* & 0.01* & -0.02* & 0.01* & 0.01* \\ \hline
\textbf{White} & -0.04* & {[}'-0.02*'{]} & -0.04* & {[}'0.03*'{]} & -0.01* & {[}'0.01*'{]} \\ \hline
\textbf{Black} & 0.02* & {[}'0.02*'{]} & 0.02* & {[}'-0.03*'{]} & 0.01* & {[}'-0.01*'{]} \\ \hline
\textbf{Age {[}14 - 52)} & -0.02* & -0.03* & 0.06* & 0.03* & 0.02* & 0.0 \\ \hline
\textbf{Age {[}52 - 67)} & 0.0 & {[}'0.03*', '-0.03*'{]} & 0.0 & {[}'-0.03*', '0.04*'{]} & -0.01* & {[}0.0, '0.0*{]} \\ \hline
\textbf{Age {[}67 - 102{]}} & 0.02* & 0.03* & -0.06* & -0.04* & -0.02* & -0.0* \\ \hline
\textbf{Overall Avg} & 0.76 (PPV) & - & 0.77 (Sensitivity) & - & 0.85 (AUROC) & - \\ \hline
\end{tabular}}
   }
\end{table*}

\begin{table*}[!h]
\centering
\floatconts{tab:Epic-auroc-std}
  {\caption{ACTFAST-Epic: AUROC with standard deviations on 150 bootstrapped samples.}}
{\scriptsize
\begin{tabular}{|c|cc|cc|}
\hline
\textbf{} & \multicolumn{2}{c|}{\textbf{Model with protected attribute}} & \multicolumn{2}{c|}{\textbf{Models without protected attributes}} \\ \hline
\textbf{Patient group} & \multicolumn{1}{c|}{\textbf{\begin{tabular}[c]{@{}c@{}}Grp - Avg  \\ AUROC\end{tabular}}} & \textbf{\begin{tabular}[c]{@{}c@{}}Grp - Avg \\ AUROC (PSM)\end{tabular}} & \multicolumn{1}{c|}{\textbf{\begin{tabular}[c]{@{}c@{}}Grp - Avg \\ AUROC\end{tabular}}} & \textbf{\begin{tabular}[c]{@{}c@{}}Grp - Avg \\ AUROC (PSM)\end{tabular}} \\ \hline
\textbf{Male} & \multicolumn{1}{c|}{-0.01*+-0.0106} & -0.01*+-0.003 & \multicolumn{1}{c|}{-0.01*+-0.0101} & -0.01*+-0.003 \\ \hline
\textbf{Female} & \multicolumn{1}{c|}{0.01*+-0.0106} & 0.01*+-0.003 & \multicolumn{1}{c|}{0.01*+-0.0101} & 0.01*+-0.003 \\ \hline
\textbf{White} & \multicolumn{1}{c|}{-0.01*+-0.0309} & {[}'0.02*+-0.003'{]} & \multicolumn{1}{c|}{-0.01*+-0.0306} & {[}'0.01*+-0.003'{]} \\ \hline
\textbf{Black} & \multicolumn{1}{c|}{0.01*+-0.0171} & {[}'-0.02*+-0.003'{]} & \multicolumn{1}{c|}{0.01*+-0.0169} & {[}'-0.01*+-0.003'{]} \\ \hline
\textbf{Age {[}14 - 52)} & \multicolumn{1}{c|}{0.02*+-0.0148} & 0.01*+-0.003 & \multicolumn{1}{c|}{0.02*+-0.0148} & -0.0+-0.003 \\ \hline
\textbf{Age {[}52 - 67)} & \multicolumn{1}{c|}{-0.01*+-0.0134} & {[}'-0.01*+-0.003', '0.0+-0.003'{]} & \multicolumn{1}{c|}{-0.01*+-0.0136} & {[}'0.0+-0.003', '0.0*+-0.003'{]} \\ \hline
\textbf{Age {[}67 - 102{]}} & \multicolumn{1}{c|}{-0.01*+-0.0142} & -0.0+-0.003 & \multicolumn{1}{c|}{-0.02*+-0.0143} & -0.0*+-0.003 \\ \hline
\textbf{Overall Avg} & \multicolumn{1}{c|}{0.85 (AUROC)} & - & \multicolumn{1}{c|}{0.85 (AUROC)} & - \\ \hline
\end{tabular}}
\end{table*}

\begin{table*}[!h]
\floatconts{tab:Epic_mistakes}
  {\caption{ACTFAST-Epic: Comparison of FNR, FPR results when the protected attributes were included and not included in the training dataset.}}
{\tiny
\begin{tabular}{|c|cccc|cccc|}
\hline
 & \multicolumn{4}{c|}{\textbf{Model with protected attributes}} & \multicolumn{4}{c|}{\textbf{Model without protected attributes}} \\ \hline
\textbf{\begin{tabular}[c]{@{}c@{}}Patient group\\ ACTFAST-Epic\end{tabular}} & \multicolumn{1}{c|}{\textbf{\begin{tabular}[c]{@{}c@{}}Grp- Avg\\ FNR\end{tabular}}} & \multicolumn{1}{c|}{\textbf{\begin{tabular}[c]{@{}c@{}}Grp - Avg\\  FNR (PSM)\end{tabular}}} & \multicolumn{1}{c|}{\textbf{\begin{tabular}[c]{@{}c@{}}Grp - Avg\\ FPR\end{tabular}}} & \textbf{\begin{tabular}[c]{@{}c@{}}Grp - Avg\\ FPR (PSM)\end{tabular}} & \multicolumn{1}{c|}{\textbf{\begin{tabular}[c]{@{}c@{}}Grp- Avg\\ FNR\end{tabular}}} & \multicolumn{1}{c|}{\textbf{\begin{tabular}[c]{@{}c@{}}Grp - Avg\\  FNR (PSM)\end{tabular}}} & \multicolumn{1}{c|}{\textbf{\begin{tabular}[c]{@{}c@{}}Grp - Avg\\ FPR\end{tabular}}} & \textbf{\begin{tabular}[c]{@{}c@{}}Grp - Avg\\ FPR (PSM)\end{tabular}} \\ \hline
\textbf{Male} & \multicolumn{1}{c|}{0.02*} & \multicolumn{1}{c|}{-0.0} & \multicolumn{1}{c|}{-0.01*} & 0.01* & \multicolumn{1}{c|}{0.01*} & \multicolumn{1}{c|}{-0.02*} & \multicolumn{1}{c|}{0.0} & 0.02* \\ \hline
\textbf{Female} & \multicolumn{1}{c|}{-0.02*} & \multicolumn{1}{c|}{0.0} & \multicolumn{1}{c|}{0.01*} & -0.01* & \multicolumn{1}{c|}{-0.01*} & \multicolumn{1}{c|}{0.02*} & \multicolumn{1}{c|}{0.0} & -0.02* \\ \hline
\textbf{White} & \multicolumn{1}{c|}{0.04*} & \multicolumn{1}{c|}{{[}-0.05*{]}} & \multicolumn{1}{c|}{0.01*} & {[}0.01*{]} & \multicolumn{1}{c|}{0.04*} & \multicolumn{1}{c|}{{[}-0.03*{]}} & \multicolumn{1}{c|}{0.0} & {[}0.02*{]} \\ \hline
\textbf{Black} & \multicolumn{1}{c|}{-0.02*} & \multicolumn{1}{c|}{{[}0.05*{]}} & \multicolumn{1}{c|}{-0.0} & {[}-0.01*{]} & \multicolumn{1}{c|}{-0.02*} & \multicolumn{1}{c|}{{[}0.03*{]}} & \multicolumn{1}{c|}{0.0} & {[}-0.02*{]} \\ \hline
\textbf{Age {[}14 - 52)} & \multicolumn{1}{c|}{-0.05*} & \multicolumn{1}{c|}{-0.03*} & \multicolumn{1}{c|}{0.01*} & 0.0* & \multicolumn{1}{c|}{-0.06*} & \multicolumn{1}{c|}{-0.03*} & \multicolumn{1}{c|}{0.02*} & 0.01* \\ \hline
\textbf{Age {[}52 - 67)} & \multicolumn{1}{c|}{0.01*} & \multicolumn{1}{c|}{{[}0.03*, -0.01*{]}} & \multicolumn{1}{c|}{0.0} & {[}-0.0*, 0.0{]} & \multicolumn{1}{c|}{0.0} & \multicolumn{1}{c|}{{[}0.03*, -0.04*{]}} & \multicolumn{1}{c|}{0.0} & {[}-0.01*, 0.02*{]} \\ \hline
\textbf{Age {[}67 - 102{]}} & \multicolumn{1}{c|}{0.04*} & \multicolumn{1}{c|}{0.01*} & \multicolumn{1}{c|}{-0.01*} & -0.0 & \multicolumn{1}{c|}{0.06*} & \multicolumn{1}{c|}{0.04*} & \multicolumn{1}{c|}{-0.02*} & -0.02* \\ \hline
\textbf{Overall Avg} & \multicolumn{1}{c|}{0.23 (FNR)} & \multicolumn{1}{c|}{-} & \multicolumn{1}{c|}{0.22 (FPR)} & - & \multicolumn{1}{c|}{0.23 (FNR)} & \multicolumn{1}{c|}{-} & \multicolumn{1}{c|}{0.22 (FPR)} & - \\ \hline
\end{tabular}

}
\end{table*}

\begin{table*}
\floatconts{tab:MIMIC-withOutsensM1}
  {\caption{Performance difference from the group average for the MIMIC-III dataset using a Random Forest classifier on a bootstrapped (150 samples) test set when the sensitive features (age, sex, and race) \textbf{were not} included in the training dataset. }}
{ \scriptsize 
\begin{tabular}{|c|c|c|c|c|c|c|}
\hline
\textbf{\begin{tabular}[c]{@{}c@{}}Patient \\ group \\ MIMIC-III\end{tabular}} & \textbf{\begin{tabular}[c]{@{}c@{}}Grp - Avg \\ PPV\end{tabular}} & \textbf{\begin{tabular}[c]{@{}c@{}}Grp -Avg \\ PPV (PSM)\end{tabular}} & \textbf{\begin{tabular}[c]{@{}c@{}}Grp - Avg \\ Sens\end{tabular}} & \textbf{\begin{tabular}[c]{@{}c@{}}Grp - Avg \\ Sens (PSM)\end{tabular}} & \textbf{\begin{tabular}[c]{@{}c@{}}Grp - Avg \\ AUROC\end{tabular}} & \textbf{\begin{tabular}[c]{@{}c@{}}Grp- Avg \\ AUROC \\ (PSM)\end{tabular}} \\ \hline
\textbf{Male} & 0.02* & 0.02* & 0.03* & 0.03* & 0.01* & 0.01* \\ \hline
\textbf{Female} & -0.02* & -0.02* & -0.03* & -0.03* & -0.01* & -0.01* \\ \hline
\textbf{Hispanic} & 0.0 & {[}'0.01*', '-0.02*'{]} & -0.01 & {[}'-0.04*', -0.0{]} & 0.0 & {[}'0.01*', -0.0{]} \\ \hline
\textbf{White} & 0.0 & {[}-0.0, '0.02*'{]} & -0.02* & {[}'-0.01*', 0.0{]} & 0.0 & {[}'-0.0*', 0.0{]} \\ \hline
\textbf{Black} & 0.0 & {[}0.0, '-0.01*'{]} & 0.03* & {[}'0.01*', '0.04*'{]} & 0.01* & {[}'0.0*', '-0.01*'{]} \\ \hline
\textbf{Age{[}18 - 57)} & -0.04* & -0.03* & -0.01* & -0.01* & -0.01* & 0.0 \\ \hline
\textbf{Age {[}57 - 74)} & 0.0 & {[}'0.03*', '-0.02*'{]} & -0.02* & {[}'0.01*', 0.0{]} & 0.01* & {[}0.0, 0.0{]} \\ \hline
\textbf{Age {[}74 - 90{]}} & 0.05* & 0.02* & 0.03* & 0.0 & 0.01* & 0.0 \\ \hline
\textbf{Overall Avg} & 0.26 (PPV) & - & 0.79 (Sensitivity) & - & 0.87 (AUROC) & - \\ \hline
\end{tabular}}
\end{table*}

\begin{table*}
\floatconts{tab:MIMIC-withsensM2}
  {\caption{Performance difference from the group average for the MIMIC-III dataset using a Random Forest classifier on a bootstrapped (150 samples) test set when the sensitive features (age, sex, and race) \textbf{were} included in the training dataset. }}
{ \scriptsize 
\begin{tabular}{|c|c|c|c|c|c|c|}
\hline
\textbf{\begin{tabular}[c]{@{}c@{}}Patient \\ group \\ MIMIC-III\end{tabular}} & \textbf{\begin{tabular}[c]{@{}c@{}}Grp - Avg \\ PPV\end{tabular}} & \textbf{\begin{tabular}[c]{@{}c@{}}Grp -Avg \\ PPV (PSM)\end{tabular}} & \textbf{\begin{tabular}[c]{@{}c@{}}Grp - Avg \\ Sens\end{tabular}} & \textbf{\begin{tabular}[c]{@{}c@{}}Grp - Avg \\ Sens (PSM)\end{tabular}} & \textbf{\begin{tabular}[c]{@{}c@{}}Grp - Avg \\ AUROC\end{tabular}} & \textbf{\begin{tabular}[c]{@{}c@{}}Grp- Avg \\ AUROC \\ (PSM)\end{tabular}} \\ \hline
\textbf{Male} & 0.02* & 0.02* & 0.02* & 0.02* & 0.01* & 0.02* \\ \hline
\textbf{Female} & -0.02* & -0.02* & -0.02* & -0.02* & -0.01* & -0.02* \\ \hline
\textbf{Hispanic} & -0.02* & {[}0.0, '-0.04*'{]} & 0.0 & {[}'0.04*', '-0.02*'{]} & 0.02* & {[}'0.03*', '0.01*'{]} \\ \hline
\textbf{White} & 0.02* & {[}'0.01*', '0.04*'{]} & -0.04* & {[}'-0.04*', '0.02*'{]} & -0.01* & {[}'-0.01*', '-0.01*'{]} \\ \hline
\textbf{Black} & 0.0 & {[}'-0.01*', -0.0{]} & 0.04* & {[}'0.04*', '-0.04*'{]} & -0.01* & {[}'0.01*', '-0.03*'{]} \\ \hline
\textbf{Age{[}18 - 57)} & -0.04* & -0.02* & 0.01* & 0.03* & -0.01* & 0.01* \\ \hline
\textbf{Age {[}57 - 74)} & -0.01* & {[}'0.02*', '-0.03*'{]} & -0.03* & {[}'-0.03*', '-0.02*'{]} & 0.0 & {[}'-0.01*', '-0.01*'{]} \\ \hline
\textbf{Age {[}74 - 90{]}} & 0.04* & 0.03* & 0.02* & 0.02* & 0.01* & 0.01* \\ \hline
\textbf{Overall Avg} & 0.26 (PPV) & - & 0.8 (Sens) & - & 0.87 (AUROC) & - \\ \hline
\end{tabular}}
\end{table*}

\end{document}